\documentclass{article}

\usepackage{natbib}
\PassOptionsToPackage{numbers}{natbib}

\usepackage[final]{nips_2018}

\usepackage{booktabs}
\usepackage{array}
\newcolumntype{x}{l}
\newcolumntype{X}{>{\scriptsize}l}
\newcolumntype{v}[1]{>{\raggedright\hspace{0pt}}p{#1}}
\newcolumntype{V}[1]{>{\scriptsize\raggedright\hspace{0pt}}p{#1}}

\usepackage[utf8]{inputenc} 
\usepackage[T1]{fontenc}    
\usepackage{hyperref}       
\usepackage{url}            
\usepackage{booktabs}       
\usepackage{amsfonts}       
\usepackage{nicefrac}       
\usepackage{microtype}      
\usepackage[page]{appendix}

\usepackage{tumcolors}
\usepackage{tummath}
\usepackage{tumunits}
\usepackage{tumabbrev}
\usepackage{subfig}
\usepackage{nth}
\usepackage[capitalise]{cleveref}

\usepackage[inline]{enumitem}
\setlist[enumerate]{itemsep=-2mm}

\newcommand{\MWeight}{\ensuremath{\M{\theta}}}

\newcommand{\VInput}{\DataVec}

\newcommand{\VHidden}{\ensuremath{\V{h}}}

\newcommand{\VCellState}{\ensuremath{\V{c}}}
\newcommand{\VForgetGate}{\ensuremath{\V{f}}}

\newcommand{\VModulationGate}{\ensuremath{\V{j}}}

\newcommand{\VInputGate}{\ensuremath{\V{i}}}

\newcommand{\VOutputGate}{\ensuremath{\V{o}}}

\usepackage{tikz}
\usetikzlibrary{fit, positioning, arrows, spy, 3d}
\usetikzlibrary{external}
\tikzsetexternalprefix{tikz/}

\usepackage{pgfplots}
\usepgfplotslibrary{groupplots}
\usepgfplotslibrary{dateplot}

\title{Convolutional LSTMs for Cloud-Robust Segmentation of Remote Sensing Imagery}

\author{
	Marc Ru\ss{}wurm \\
	Chair of Remote Sensing Technology\\
	Technical University of Munich \\
	\texttt{marc.russwurm@tum.de} \\
	\And
	Marco Körner \\
	Chair of Remote Sensing Technology \\
	Technical University of Munich \\
	\texttt{marco.koerner@tum.de} \\
}

\begin{document}
	
	\maketitle
	
	\begin{abstract}
        Clouds frequently cover the Earth's surface and pose an omnipresent challenge to optical Earth observation methods.
		The vast majority of remote sensing approaches either selectively choose single cloud-free observations or employ a pre-classification strategy to identify and mask cloudy pixels.
		We follow a different strategy and treat cloud coverage as noise that is inherent to the observed satellite data.
		In prior work, we directly employed a straightforward \emph{convolutional long short-term memory} network for vegetation classification without explicit cloud filtering and achieved state-of-the-art classification accuracies.
		In this work, we investigate this cloud-robustness further by visualizing internal cell activations and performing an ablation experiment on datasets of different cloud coverage.
		In the visualizations of network states, we identified some cells in which modulation and input gates closed on cloudy pixels. 
		This indicates that the network has internalized a cloud-filtering mechanism without being specifically trained on cloud labels.
		Overall, our results question the necessity of sophisticated pre-processing pipelines for multi-temporal deep learning approaches.
		We have published the source code of your work on GitHub\footnote{\url{https://github.com/TUM-LMF/MTLCC}}.
	\end{abstract}
		
	\vspace{-1em}
	\section{Introduction}
	
	An increasing number of satellites monitor dynamic spatiotemporal processes on the Earth's surface.
	Optical satellites measure spectral reflectances at multiple short electromagnetic wavelengths in regular time intervals of a few days.
	The surface, however, is often completely or partially covered by clouds. 
	This poses an omnipresent challenge for the majority of remote sensing approaches are designed with cloud-free imagery in mind.
	In prior published work \citep{russwurm2018multi, russwurm2017temporal}, we focused on the identification of crop classes with convolutional recurrent neural networks \citep{xingjian2015convolutional} to approximate life cycle events of vegetation.  
	These phenological cycles and other land cover dynamics can be monitored at weekly intervals at spatial resolutions of several meters which allows distinguishing large single objects.
	We observed that state-of-the-art accuracies could be achieved without dedicated pre-processing mechanisms.
	In this work, we investigate this cloud-robustness further and treat clouds as data-inherent noise.
	
	\section{Related Work}
	
	Clouds distinguish themselves from ground pixels by their high reflectance.
	Rule-based models \citep{hollstein2016ready, zhu2012object, hagolle2010multi} are used for many remote sensing applications and rely on expert-designed features \citep{Li18clouds}.
	The \texttt{fmask} algorithm \citep{zhu2012object} and improved versions \citep{zhu2015improvement, frantz2015enhancing} implement a projection of the detected cloud on the surface to additionally predict the shadow cast by the cloud.
	Other approaches extract multi-temporal features and utilize the sudden temporal increase in reflectance \citep{hagolle2010multi}.
	\emph{Convolutional neural networks (CNNs)} have also shown to compare well on cloud and cloud-shadow classification \citep{Li18clouds} indicating that these features can be learned from the data by deep neural networks.
	%
	However, masking single pixels by a pre-classification introduces an additional layer of complexity and raises the question of how to treat these masked pixels accordingly.
	For instance, one recently published approach \citep{interdonato2018duplo} replaces cloudy pixels by temporal interpolation of cloud-free observations, while we introduced pre-classified cloud labels as additional prediction class in early work on this topic \citep{russwurm2017temporal}.
	Overall, cloud-filtering remains a pre-processing necessity for most remote sensing approaches that are prone to fail in the presence of data noise.
	Similar to our work, only a few approaches have tried to design robust methods that do not require this additional pre-classification step. 
	For instance, \citet{man2018improvement} used ensemble-based methods of supervised classifiers which have shown a certain robustness to the presence of clouds.

	\vspace{-.5em}
	\section{Method}
	
	In this section, we outline the theoretical basis of \emph{convolutional long short-term memory (convLSTM)} networks utilized in this work and provide detail on the employed network topology.
	
	\begin{figure}
		\includegraphics[width=\textwidth]{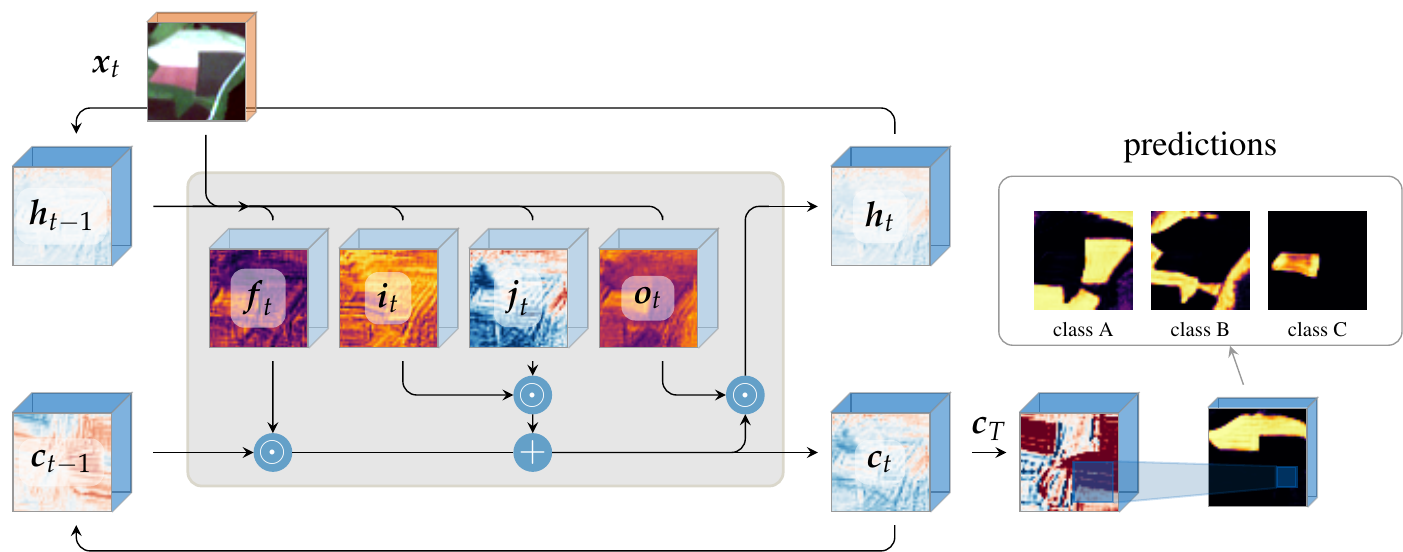}
		\caption{Illustration of the two-component convolutional long short-term memory network (LSTM) topology employed in this work. %
			Each input image $\VInput_t$ of a series of $T$ images is passed sequentially to the LSTM encoder. Classification relevant features are extracted to the internal cell state tensor $\VCellState_T$.
			A second convolutional layer compresses the dimensionality to the number of classes which yields activations per class.}
		\label{fig:network}
	\end{figure}
	
	\subsection{Convolutional Long Short-term Memory}
	
	\emph{Long short-term memory (LSTM)} networks \citep{hochreiter1997long} implement internal gates to control the gradient-flow through time and an additional container for long-term memory $\VCellState_t$.
	This yields the LSTM update $(\VHidden_t, \VCellState_{t-1}) \leftarrow (\VInput_t, \VHidden_{t-1}, \VCellState_{t-1})$ that
	maps an input $\VInput_t$ and short-term context $\VHidden_{t-1}$ to a hidden representation $\VHidden_t$. 
	Additionally, a long-term cell state $\VCellState_{t-1}$ is updated to $\VCellState_t$ at each iteration and can store information for a theoretically unlimited number of iteration.
	Three gates control the update of the cell state
	$\VCellState_t \leftarrow \VCellState_{t-1} \odot \VForgetGate_t + \VInputGate_t \odot \VModulationGate_t
	$
	by element-wise multiplication $\odot$. 
	The forget gate
	$
	\VForgetGate_t = \sigma \left( \conv{\VInput_t}{\MWeight_\text{fx}} + \conv{\VHidden_{t-1}}{\MWeight_\text{fh}} + \V{1}\right)
	$
	evaluates the influence of the previous cell state $\VCellState_{t-1}$ with a sigmoidal $\sigma \left(\cdot\right) \in \left]0,1\right[$ activation function.
	The input and modulation gates 
	\begin{equation}
	\label{eq:inputgate}
	\VInputGate_t = \sigma \left( \conv{\VInput_t}{\MWeight_\text{ix}} + \conv{\VHidden_{t-1}}{\MWeight_\text{ih}} \right) \text{, and}\quad  \VModulationGate_t = \tanh \left( \conv{\VInput_t}{\MWeight_\text{jx}} + \conv{\VHidden_{t-1}}{\MWeight_\text{jh}} \right)
	\end{equation}
	are element-wise multiplied for the cell state update.
	The output gate \\
	$
	\VOutputGate_t = \tanh \left( \conv{\VInput_t}{\MWeight_\text{ox}} + \conv{\VHidden_{t-1}}{\MWeight_\text{oh}} \right)
	$
	determines together with the cell state the current cell output \\ $\VHidden_t \leftarrow \VOutputGate_t \odot \VCellState_t$.
	Convolutional recurrent networks implement a convolution, denoted by $\conv{}{}$, instead of a matrix multiplication.
	Each respective gate activation, referred by subscripts $\text{f}, \text{i}, \text{j}, \text{o}$, is controlled by trainable weights for input $\MWeight_\text{fx}, \MWeight_\text{ix}, \MWeight_\text{jx}, \MWeight_\text{ox} \in \R^{k \times k \times d  \times r}$ and hidden representation $\MWeight_\text{fh}, \MWeight_\text{ih}, \MWeight_\text{jh}, \MWeight_\text{oh} \in \R^{k \times k \times r  \times r}$ where $d$ represents the number of image channels, $k$ the convolutional kernel size, and $r$ a hyper-parameter that determins the number of the hidden states in the recurrent layer.
	With this change, image data of certain width, height and depth can be processed with convolutions partially connecting the local pixel neighborhoods between layers.
	
	\subsection{Network Topology}
	
	We utilize this single-layer convolutional LSTM neural network to encode a sequence of $T$ satellite images to the fixed length representation $\VCellState_T$, as illustrated in \cref{fig:network}.
	In our initial work \citep{russwurm2018multi}, we found that 256 recurrent cells performed best and used this hyper-parameter for the number of hidden states within the ConvLSTM network. 
	To balance the influence of the sequence order, we also encode the reversed sequence and append the final cell states.
	Applying softmax normalization produces activations that can be interpreted as network-confidences per class.
	We used convolutional kernels of $3 \times 3px$ throughout the network. 
	To train, we evaluate the cross-entropy between the last layer and a one-hot representation of the ground truth labels.
	The influence of each weight on the evaluated loss is determined by back-propagated gradients which are used to iteratively adjustment the respective weight using the Adam optimizer \citep{kingma2014adam} with a learning rate of $\lambda = 0.001$.
	
	\begin{figure}
		\includegraphics[width=\textwidth]{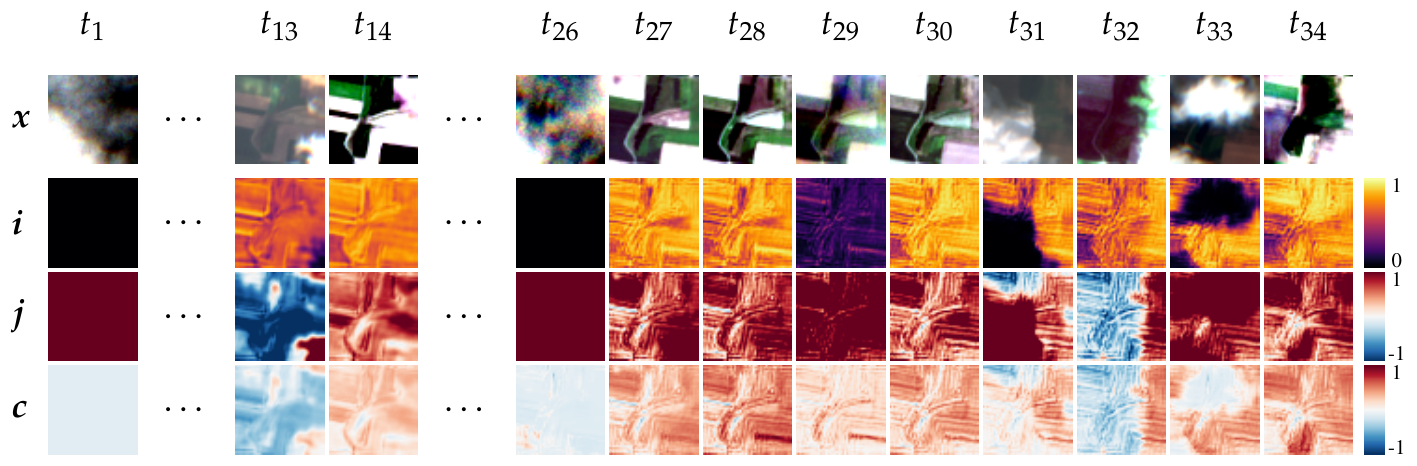}
		
		\caption{%
			Activations of the cell state and selected gates of one convolutional LSTM cell that indicate that the cell has internalized a cloud-fiiltering scheme.
			The input gate $\VInputGate$ in this specific cell seems to be assigned values of zero on cloudy pixels as seen at steps $t=\{13,26,31,33\}$.     
		}
		\label{fig:activations}
	\end{figure}
	
	\vspace{-.5em}
	\section{Results}
	
	The primary objective for this network was to identify the type of cultivated crops in an area of interest of \SI{100}{\kilo\meter} $\times$ \SI{40}{\kilo\meter}, as described at greater detail in \citet{russwurm2018multi}.
	Hence, we trained our network end-to-end on label data describing the crop-type on distinct field parcels and
	provided no additional label information about cloud coverages.
	We used a sequence of 46 \textsc{Sentinel 2} satellite images from the year 2016 for this objective.
	This satellite measures the reflectances of 13 spectral bands at \SI{10}{\meter}, \SI{20}{\meter}, and \SI{60}{\meter} resolution.    
	To harmonize the data sources, we bi-linearly interpolated these to \SI{10}{\meter} resolution and rasterized the crop labels accordingly.
	Additional information about area of interest and data partitioning strategy is given in Appendix \ref{app:data}. 
	In the remainder of this section, we evaluate the robustness of the proposed network to cloud coverage.
	
	\subsection{Exploration of Hidden LSTM States}
	\label{sec:activations}
	
	We trained the network on field crop labels for thirty epochs using raw sequences of cloudy and non-cloudy observations.
	The top row of \cref{fig:activations} shows an particular example sequence with $48 \times 48px$ images.
	The following rows illustrate activations of the internal convolutional LSTM for input gate $\VInputGate$, modulation gate $\VModulationGate$, and cell state $\VCellState$ at each input time $t$.
	While all of the 256 recurrent cells likely contribute to the classification decision, only a few were visually interpretable similar to the shown example, as can be compared in Appendix \ref{app:activations}.
	The activations in \cref{fig:activations} of these gates in the second and third row show that the input gate $\VInputGate$ approaches zero at pixels that are covered by clouds.
	This effect can be observed at time steps $t=\{13,26,31,33\}$.
	At time $t=32$ the input gate seems unchanged, however, the modulation gate $\VModulationGate$ changes sign.
	Overall, these results indicate that the convolutional recurrent network has internalized a mechanism for cloud-filtering. 
	
	\subsection{Ablation Experiment on Cloud Coverage}
	\label{sec:robustness}
	
\begin{figure}
		\includegraphics[width=\textwidth]{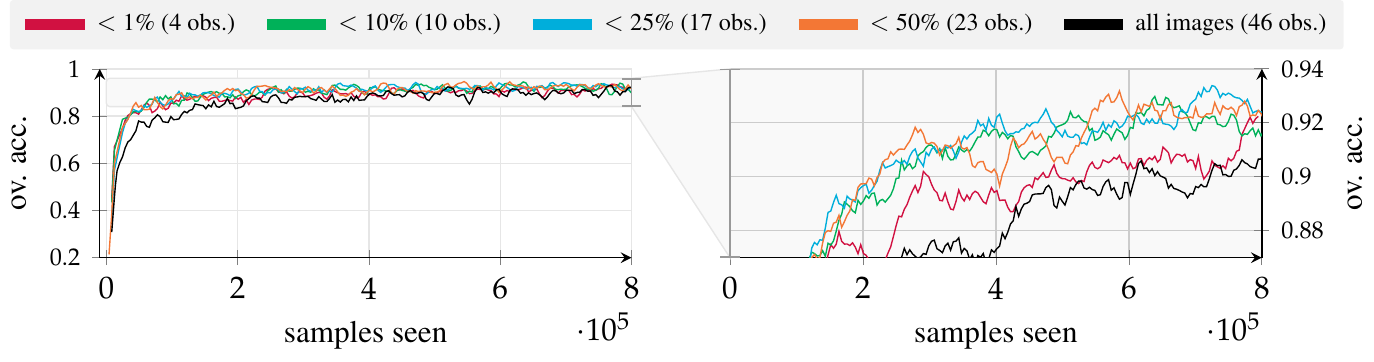}
		\caption{%
			Overall accuracy over the training progress recorded on the validation partition with different degrees of cloud coverage. The network achieves similar accuracies on cloudy and non-cloudy data.
		}
		\label{fig:cloudoa}
	\end{figure}

	In this experiment, we trained the network on datasets with different degrees of cloud coverage.
	To determine the actual coverage of clouds per observation, all satellite images have been pre-processed using the \texttt{fmask} algorithm implemented in the \texttt{Sen2Cor} software, as is common practice in remote sensing \citep{conrad2014,Foerster2012}. 
	With this, a cloud coverage pixel ratio per observation can be calculated.
	Based on this, several sub-datasets have been created with either all 46 observations, the 26 images covered with less than \SI{50}{\percent}, 17 images with less than \SI{25}{\percent}, 10 with less than \SI{10}{\percent}, and 4 completely cloud-free images. 
	We trained the network on these pre-filtered datasets and show the overall accuracy recorded during training on validation data in \cref{fig:cloudoa}.
	Even though the ratio of cloudy images varies greatly between these datasets, the classification accuracy remains similar for all of the sub-sampled datasets.
	The right graph shows a zoomed view and reveals some differences between the dataset performances.
	Datasets containing all observations and the four completely cloud-free observations have been slightly worse classified than the intermediate ones of \SI{10}{\percent}, \SI{25}{\percent}, and \SI{50}{\percent} coverage.
	It seems that the manual removal of the completely cloud-covered observations had a minor benefit on the classification accuracy.
	Also, the four cloud-free observations were classified slightly worse which indicates that these may have missed some characteristic vegetation-related events.
	Intuitively, these results show a trade-off between restrictions on cloud coverage and sequence length and demonstrate that cherry-picking single cloud-free observations may lead to inferior classification accuracy.
	Overall, these results demonstrate the robustness of the convolutional long short-term memory network to handle data containing temporal noise induced by cloud coverage.
	
	\vspace{-.5em}
	\section{Conclusion}
	
	Noise in temporal data is a common challenge for a variety of disciplines.
	In this work, we focused on noise induced by cloud coverage in multi-temporal remote sensing imagery.
	Most Earth observation approaches either select few completely cloud-free observations or use a pre-classification to mask cloudy pixels.
	The experiments of this work showed that this cloud-induced temporal noise can be suppressed by	 training on an other objective in an end-to-end fashion with an appropriate model design.
	Our results demonstrate that convolutional recurrent networks are able to consistently extract the classification-relevant features from observations between clouds without dedicated cloud labels. 
	Overall, this work questions the necessity of sophisticated, partly hand-crafted pre-processing pipelines for remote sensing imagery.

	\section*{Revisions}
	
	{
	\small
	\begin{description}[itemsep=0mm]
		\item[v1] submitted version to the NeurIPS Spatiotemporal Workshop~ \href{https://openreview.net/references/pdf?id=SkMdGcCYQ}{OpenReview v1}
		\item[v2] incorporated first OpenReview comments and made minor text changes. \href{https://arxiv.org/abs/1811.02471v1}{arXiv:1811.02471v1}  
		\item[v3] major text refinement and added appendix to paper. Workshop cameraready version \href{https://openreview.net/references/pdf?id=HySzgr1kV}{OpenReview v2} 
		\item[v4] changed NIPS to NeurIPS in the first page. otherwise identical to v3. \href{https://arxiv.org/abs/1811.02471v2}{arXiv:1811.02471v2} 
	\end{description}
	}

	{
		\setlength{\bibsep}{1pt plus 0.5ex}
		\bibliographystyle{plainnat}
		\bibliography{references}

\begin{thebibliography}{15}
\providecommand{\natexlab}[1]{#1}
\providecommand{\url}[1]{\texttt{#1}}
\expandafter\ifx\csname urlstyle\endcsname\relax
  \providecommand{\doi}[1]{doi: #1}\else
  \providecommand{\doi}{doi: \begingroup \urlstyle{rm}\Url}\fi

\bibitem[Conrad et~al.(2014)Conrad, Dech, Dubovyk, Fritsch, Klein, L{\"{o}}w,
  Schorcht, and Zeidler]{conrad2014}
Christopher Conrad, Stefan Dech, Olena Dubovyk, Sebastian Fritsch, Doris Klein,
  Fabian L{\"{o}}w, Gunther Schorcht, and Julian Zeidler.
\newblock {Derivation of temporal windows for accurate crop discrimination in
  heterogeneous croplands of Uzbekistan using multitemporal RapidEye images}.
\newblock \emph{Computers and Electronics in Agriculture}, 103:\penalty0
  63--74, 2014.
\newblock ISSN 01681699.
\newblock \doi{10.1016/j.compag.2014.02.003}.
\newblock URL \url{http://dx.doi.org/10.1016/j.compag.2014.02.003}.

\bibitem[Foerster et~al.(2012)Foerster, Kaden, Foerster, and
  Itzerott]{Foerster2012}
Saskia Foerster, Klaus Kaden, Michael Foerster, and Sibylle Itzerott.
\newblock {Crop type mapping using spectral–temporal profiles and
  phenological information}.
\newblock \emph{Computers and Electronics in Agriculture}, 89:\penalty0 30--40,
  2012.
\newblock ISSN 01681699.
\newblock \doi{10.1016/j.compag.2012.07.015}.
\newblock URL
  \url{http://linkinghub.elsevier.com/retrieve/pii/S0168169912002013}.

\bibitem[Frantz et~al.(2015)Frantz, R{\"o}der, Udelhoven, and
  Schmidt]{frantz2015enhancing}
David Frantz, Achim R{\"o}der, Thomas Udelhoven, and Michael Schmidt.
\newblock Enhancing the detectability of clouds and their shadows in
  multitemporal dryland landsat imagery: extending fmask.
\newblock \emph{IEEE Geoscience and Remote Sensing Letters}, 12\penalty0
  (6):\penalty0 1242--1246, 2015.

\bibitem[Hagolle et~al.(2010)Hagolle, Huc, Pascual, and
  Dedieu]{hagolle2010multi}
Olivier Hagolle, Mireille Huc, D~Villa Pascual, and G{\'e}rard Dedieu.
\newblock A multi-temporal method for cloud detection, applied to formosat-2,
  ven$\mu$s, landsat and sentinel-2 images.
\newblock \emph{Remote Sensing of Environment}, 114\penalty0 (8):\penalty0
  1747--1755, 2010.

\bibitem[Hochreiter and Schmidhuber(1997)]{hochreiter1997long}
Sepp Hochreiter and J{\"u}rgen Schmidhuber.
\newblock Long short-term memory.
\newblock \emph{Neural computation}, 9\penalty0 (8):\penalty0 1735--1780, 1997.

\bibitem[Hollstein et~al.(2016)Hollstein, Segl, Guanter, Brell, and
  Enesco]{hollstein2016ready}
Andr{\'e} Hollstein, Karl Segl, Luis Guanter, Maximilian Brell, and Marta
  Enesco.
\newblock Ready-to-use methods for the detection of clouds, cirrus, snow,
  shadow, water and clear sky pixels in sentinel-2 msi images.
\newblock \emph{Remote Sensing}, 8\penalty0 (8):\penalty0 666, 2016.

\bibitem[Interdonato et~al.(2018)Interdonato, Ienco, Gaetano, and
  Ose]{interdonato2018duplo}
Roberto Interdonato, Dino Ienco, Raffaele Gaetano, and Kenji Ose.
\newblock Duplo: A dual view point deep learning architecture for time series
  classification.
\newblock \emph{arXiv preprint arXiv:1809.07589}, 2018.

\bibitem[Kingma and Ba(2014)]{kingma2014adam}
Diederik~P Kingma and Jimmy Ba.
\newblock Adam: A method for stochastic optimization.
\newblock \emph{arXiv preprint arXiv:1412.6980}, 2014.

\bibitem[Li et~al.(2018)Li, Shen, Cheng, Liu, You, and He]{Li18clouds}
Zhiwei Li, Huanfeng Shen, Qing Cheng, Yuhao Liu, Shucheng You, and Zongyi He.
\newblock Deep learning based cloud detection for remote sensing images by the
  fusion of multi-scale convolutional features.
\newblock \emph{arXiv preprint arXiv:1810.05801}, 2018.

\bibitem[Man et~al.(2018)Man, Nguyen, Bui, Lasko, and
  Nguyen]{man2018improvement}
Chuc~Duc Man, Thuy~Thanh Nguyen, Hung~Quang Bui, Kristofer Lasko, and Thanh
  Nhat~Thi Nguyen.
\newblock Improvement of land-cover classification over frequently
  cloud-covered areas using landsat 8 time-series composites and an ensemble of
  supervised classifiers.
\newblock \emph{International Journal of Remote Sensing}, 39\penalty0
  (4):\penalty0 1243--1255, 2018.

\bibitem[Ru{\ss}wurm and K{\"o}rner(2017)]{russwurm2017temporal}
Marc Ru{\ss}wurm and Marco K{\"o}rner.
\newblock Temporal vegetation modelling using long short-term memory networks
  for crop identification from medium-resolution multi-spectral satellite
  images.
\newblock In \emph{CVPR Workshops}, pages 1496--1504, 2017.

\bibitem[Ru{\ss}wurm and K{\"o}rner(2018)]{russwurm2018multi}
Marc Ru{\ss}wurm and Marco K{\"o}rner.
\newblock Multi-temporal land cover classification with sequential recurrent
  encoders.
\newblock \emph{ISPRS International Journal of Geo-Information}, 7\penalty0
  (4):\penalty0 129, 2018.

\bibitem[Xingjian et~al.(2015)Xingjian, Chen, Wang, Yeung, Wong, and
  Woo]{xingjian2015convolutional}
Shi Xingjian, Zhourong Chen, Hao Wang, Dit-Yan Yeung, Wai-Kin Wong, and
  Wang-chun Woo.
\newblock Convolutional lstm network: A machine learning approach for
  precipitation nowcasting.
\newblock In \emph{Advances in neural information processing systems}, pages
  802--810, 2015.

\bibitem[Zhu and Woodcock(2012)]{zhu2012object}
Zhe Zhu and Curtis~E Woodcock.
\newblock Object-based cloud and cloud shadow detection in landsat imagery.
\newblock \emph{Remote sensing of environment}, 118:\penalty0 83--94, 2012.

\bibitem[Zhu et~al.(2015)Zhu, Wang, and Woodcock]{zhu2015improvement}
Zhe Zhu, Shixiong Wang, and Curtis~E Woodcock.
\newblock Improvement and expansion of the fmask algorithm: Cloud, cloud
  shadow, and snow detection for landsats 4--7, 8, and sentinel 2 images.
\newblock \emph{Remote Sensing of Environment}, 159:\penalty0 269--277, 2015.

\end{thebibliography}
	}

%
%
%
	\newpage
	
	\begin{appendices}
		
		This addendum provides further background information on the crop classification task out of which the presented evaluation on cloud-robustness has emerged.
		
		\section{Dataset and Area of Interest}
		\label{app:data}
		
\begin{figure}[h]
			\includegraphics{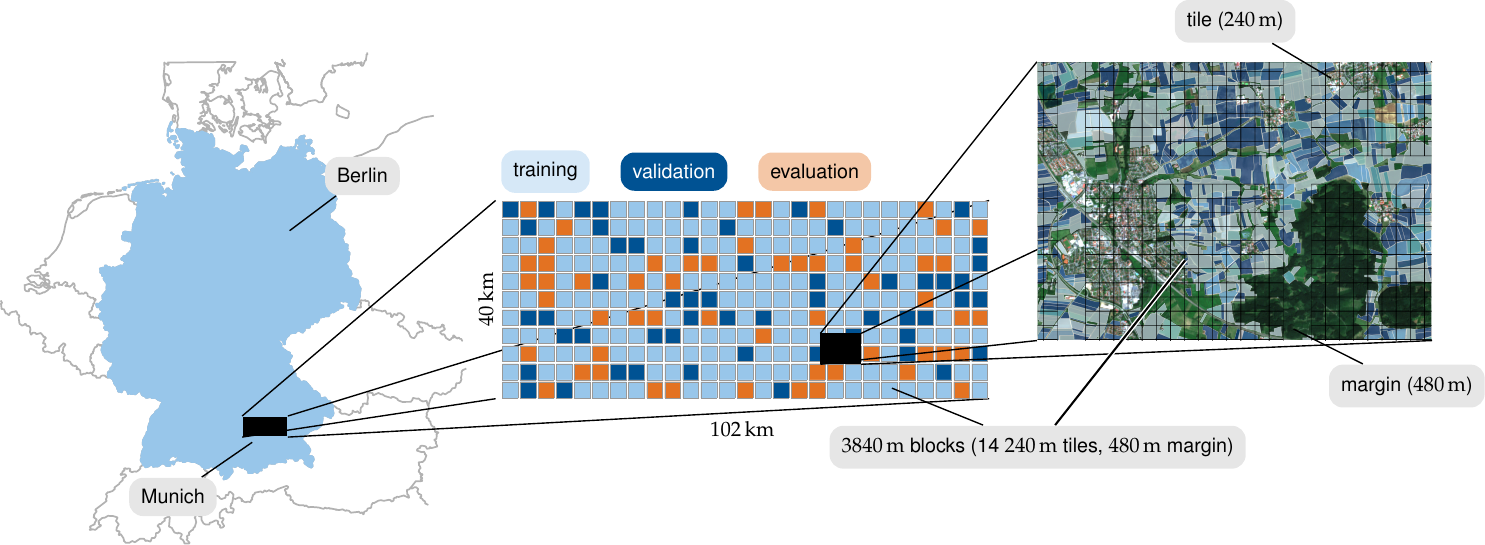}
			\caption{Overview over the area of interest (AOI)}
			\label{fig:aoi}
		\end{figure}

\begin{figure}[h]
			
			\includegraphics{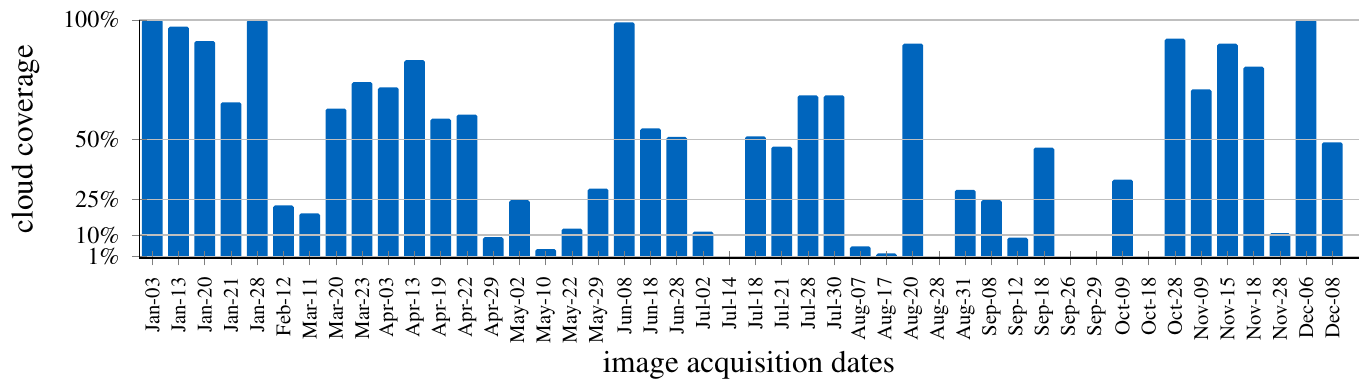}
			
			\caption{Actual cloud coverage over the area of interest obtained from the \texttt{Sen2cor} process for atmospheric correction.}
			\label{fig:scl}
		\end{figure}

		The area of interest (AOI) north of Munich, Germany, was divided into partitions for the training, validation, and evaluation, as shown in \cref{fig:aoi}.
		This ensures independence between the optimization of network parameters, hyper-parameter tuning, and final evaluation. 
		Further, we took special care to avoid introducing dependence between these partitions by spatial proximity by subdividing the AOI into rectangular blocks of 3.84 km $\times$ 3.84 km (multiples of the chosen tile sizes 240 m and 480 m). Margins between blocks of 480m avoid that tiles of two distinct partitions are located in immediate proximity. 
		These blocks were randomly assigned to the partitions in a 4:1:1 train/valid/eval ratio. 
		All tiles (240m and 480m) that were fully within the respective blocks have been assigned to their respective partition. 
		The overall cloud coverage over the area of interest is shown in \cref{fig:scl}.
		
		\vspace{2em}

		\vspace{2em}
		
		\newpage
		\section{ConvLSTM Cell Activations}
		\label{app:activations}
		
		Further internal cell activation using the same network with 256 hidden states on two tiles. The top row shows the input image in an RGB representation. The gate activations of five hidden subsequent states are shown below. Hidden state $47$ (denoted by raised index) shows a sensitivity to cloud coverage in both tiles. 
		
		\subsection{Tile A}
		\includegraphics{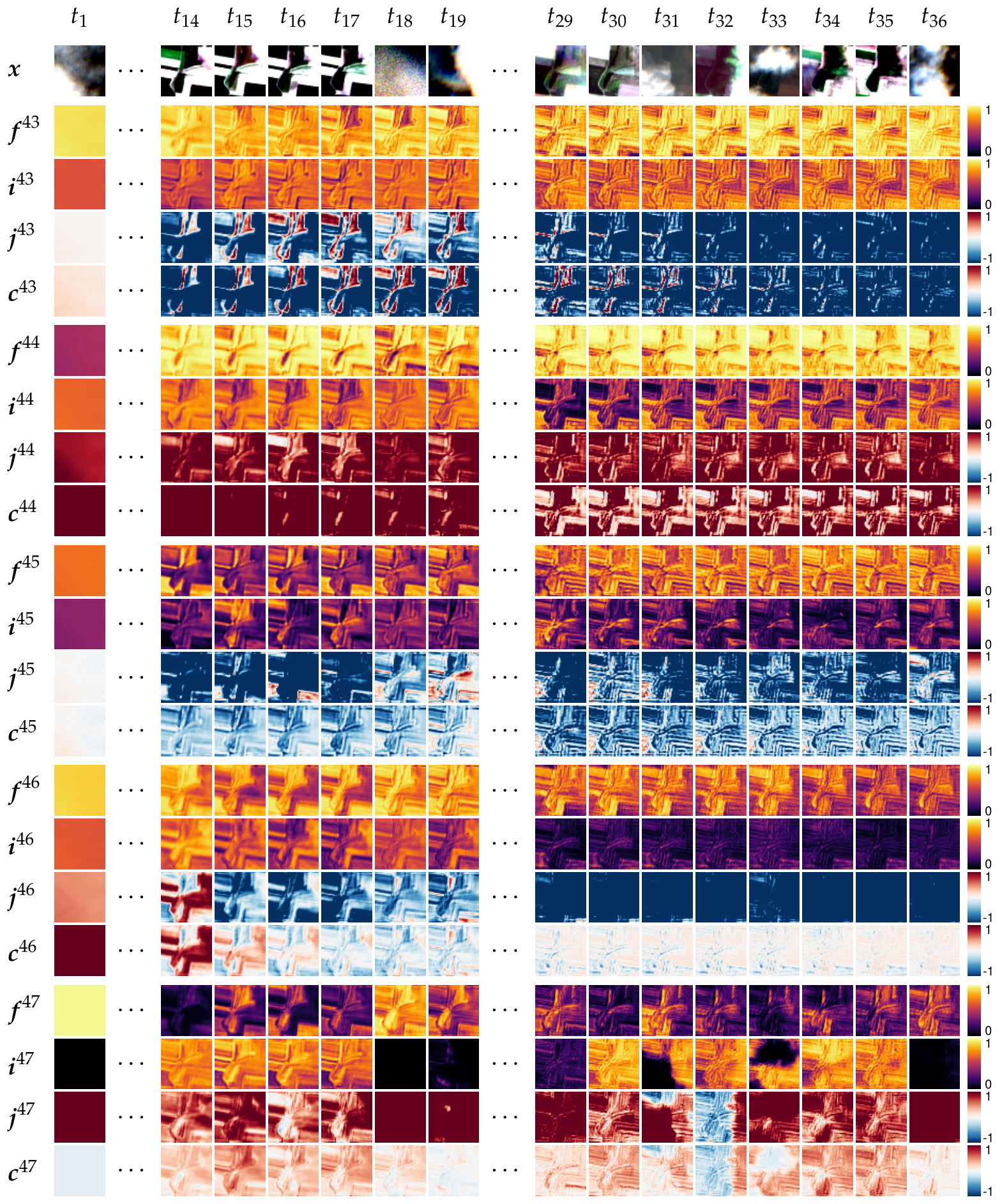}
		\newpage
		\subsection{Tile B}
		\includegraphics{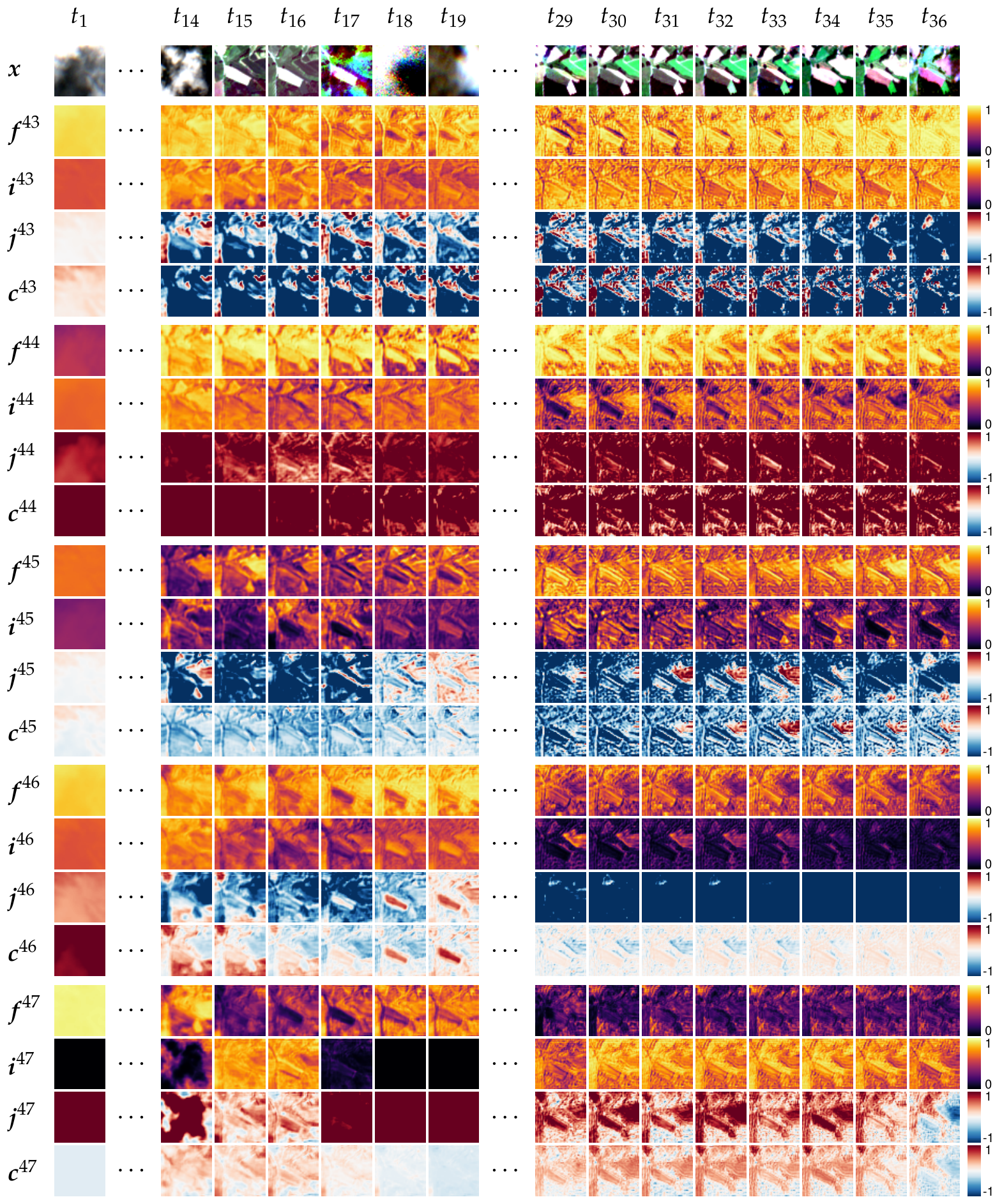}
		
	\end{appendices}
	
\end{document}